\documentclass{article}
\usepackage[preprint]{spconf,amsmath,graphicx}

\copyrightnotice{\copyright2022 IEEE}
\toappear{Appeared in {\it Proc.\ ICASSP2022, May 23-27, 2022, Singapore}: https://ieeexplore.ieee.org/document/9746641}

\usepackage[utf8]{inputenc} 
\usepackage[T1]{fontenc}    
\usepackage[hyperfootnotes=false]{hyperref}       
\usepackage{url}            
\usepackage{booktabs}       
\usepackage{amsfonts}       
\usepackage{nicefrac}       
\usepackage{microtype}      
\usepackage{xcolor}         

\usepackage{float}
\usepackage{amsmath}
\usepackage{mathtools}
\usepackage{array}
\usepackage{textcomp}
\usepackage{caption}

\newcommand\blfootnote[1]{%
  \begingroup
  \renewcommand\thefootnote{}\footnote{#1}%
  \addtocounter{footnote}{-1}%
  \endgroup
}

\title{IMPORTANCE SAMPLING CAMS FOR WEAKLY-SUPERVISED SEGMENTATION}

\name{Arvi Jonnarth\sthanks{Affiliation: Husqvarna Group, Huskvarna, Sweden.} \qquad Michael Felsberg\sthanks{Co-affiliation: the University of KwaZulu-Natal, Durban, South Africa.}}
\address{Linköping University, Sweden}

\begin{document}
\ninept
\maketitle
\begin{abstract}
Classification networks can be used to localize and segment objects in images by means of class activation maps (CAMs). However, without pixel-level annotations, classification networks are known to (1) mainly focus on discriminative regions, and (2) to produce diffuse CAMs without well-defined prediction contours. In this work, we approach both problems with two contributions for improving CAM learning. First, we incorporate \textit{importance sampling} based on the class-wise probability mass function induced by the CAMs to produce stochastic image-level class predictions. This results in CAMs which activate over a larger extent of objects. Second, we formulate a \textit{feature similarity loss term} which aims to match the prediction contours with edges in the image. As a third contribution, we conduct experiments on the PASCAL VOC 2012 benchmark dataset to demonstrate that these modifications significantly increase the performance in terms of contour accuracy, while being comparable to current state-of-the-art methods in terms of region similarity.\blfootnote{This work was partially supported by the Wallenberg AI, Autonomous Systems and Software Program (WASP) funded by the KAW foundation, and SNIC, partially funded by VR through grant agreement no.~2018-05973.}
\end{abstract}


\section{Introduction}
\label{sec_introduction}

The great advancements of deep learning methods in recent years have had a large impact on many computer vision tasks, with no exception for semantic segmentation. The ability to automatically segment images has been found useful in many applications, including autonomous driving~\cite{cordts2016cvpr}, video surveillance~\cite{gruosso2021mta} and medical image analysis~\cite{minaee2021tpami}. Fully-supervised segmentation frameworks have achieved remarkable results by utilizing large datasets of pixel-wise annotated images. However, these annotations require a significant manual labelling effort which increases with the dataset size. Image-level weakly-supervised semantic segmentation (WSSS) aims to alleviate the labelling effort required in the fully supervised case. Instead of requiring human-annotated pixel-wise segmentation masks, the only source of supervision are image-level classification labels. This opens the possibility to train segmentation models on existing large-scale datasets where pixel-level labelling is infeasible to acquire.

\begin{figure}[t]
	\setlength\abovecaptionskip{-5pt}
	\begin{center}
	\addtolength{\leftskip} {-2cm}
	\addtolength{\rightskip}{-2cm}
	\begin{tabular}{c>{\hspace{-10pt}}c>{\hspace{-10pt}}c>{\hspace{-10pt}}c>{\hspace{-10pt}}c>{\hspace{-10pt}}c}
		\includegraphics[width=.16\linewidth]{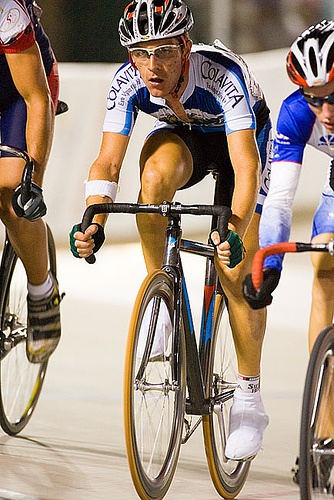} &
		\includegraphics[width=.16\linewidth]{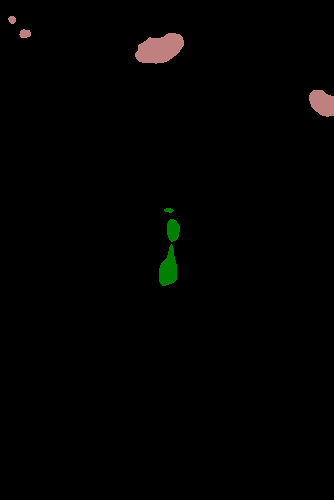} &
		\includegraphics[width=.16\linewidth]{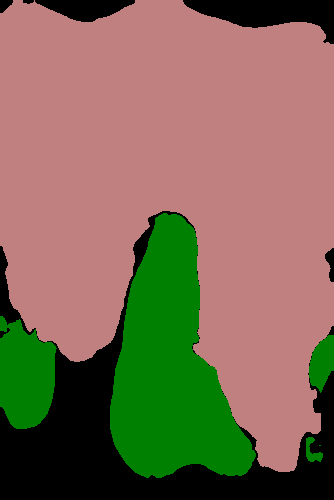} &
		\includegraphics[width=.16\linewidth]{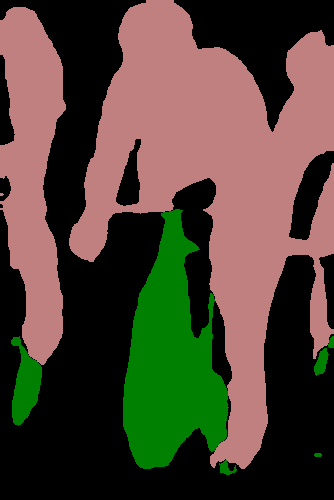} &
		\includegraphics[width=.16\linewidth]{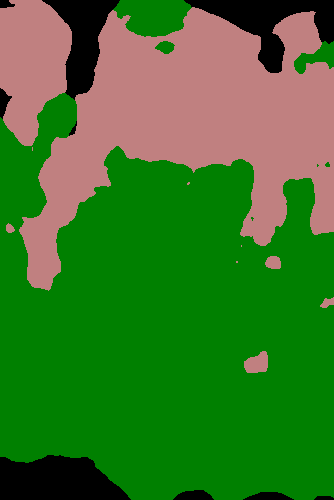} &
		\includegraphics[width=.16\linewidth]{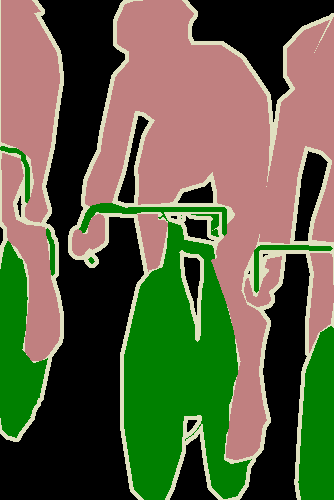} \\
		(a) & (b) & (c) & (d) & (e) & (f)
	\end{tabular}
	\end{center}
	\caption{CAM comparison. (a) Input image; pseudo-labels with (b)~max pooling, (c) importance sampling, (d) importance sampling and feature similarity loss, and (e) SEAM~\cite{wang2020cvpr}; (f) ground truth.}
	\label{fig_comparison}
\end{figure}

A common approach to WSSS is to first train a classification network with global average pooling (GAP) to produce class activation maps (CAMs)~\cite{zhou2016cvpr}. The CAMs are used to generate pseudo-labels for supervising the final segmentation network~\cite{wang2020cvpr, ahn2018cvpr, lee2019cvpr}. However, classification networks are known to (1) mainly focus on discriminative regions as opposed to the whole extent of objects, and (2) to produce overly smooth and diffuse CAMs without well-defined prediction contours~\cite{zhou2016cvpr, ahn2018cvpr, lee2019cvpr}. In this work, we improve the CAMs in these two aspects by substituting GAP with \textit{importance sampling} based on the class-wise probability mass function induced by the CAMs to produce stochastic image-level class predictions during training. This leads to CAMs which activate over a larger extent of the objects and do not only focus on the discriminative regions. Additionally, we formulate a new \textit{feature similarity loss term} which aims to match the prediction contours with edges in the image. We show experimentally that this loss term significantly increases the contour accuracy compared to our baseline method SEAM~\cite{wang2020cvpr}. Figure \ref{fig_comparison} demonstrates the benefits of our contributions by comparing pseudo-labels generated from CAMs. Finally, we perform experiments on the PASCAL VOC 2012 benchmark dataset to demonstrate that our method is comparable to current state-of-the-art WSSS methods in terms of region similarity.

\section{Related work}

Weakly-supervised semantic segmentation touches upon several different topics, for which we describe the related work below.

\textbf{Types of weak supervision.} Depending on the application and available data, one might have access to different types of annotations. Therefore, it is of interest to find methods which can learn to segment images based on various types of weak supervision. Semantic segmentation networks have successfully been trained in the past without access to full annotations in the form of segmentation masks. This has been achieved by utilizing different types of weak labels during training, such as bounding boxes~\cite{dai2015iccv, khoreva2017cvpr, papandreou2015iccv}, scribbles~\cite{lin2016cvpr, vernaza2017cvpr}, points~\cite{bearman2016eccv} and classification labels, in decreasing levels of supervision. Classification labels have been the most popular as they require the least amount of manual labelling effort, and it is clear that they can be gathered from any of the stronger label types. Classification labels have been used for learning CAMs~\cite{wang2020cvpr, lee2019cvpr} and pixel affinities~\cite{ahn2018cvpr}, constraining the neural network output~\cite{pathak2015iccv}, seeding, expanding and constraining the predictions~\cite{kolesnikov2016eccv}, as well as using them in combination with saliency maps~\cite{wei2016tpami}, all in order to increase the quality of the segmentation predictions. In this work, we focus on the lowest level of weak supervision, that is, image-level classification labels.

\textbf{Class activation maps.} Class activation maps~\cite{zhou2016cvpr, selvaraju2017iccv} have been used to visualize what part of the input a neural network uses to base its prediction on. CAMs are usually produced using GAP in the final layers, where class predictions are propagated into pixel space. Due to their ability to give an indication of the position and size of objects without ground-truth segmentation masks, they have been adapted in many WSSS methods. Ahn et al.~\cite{ahn2018cvpr} use CAMs as a starting point for predicting pixel affinities which they use to get a better segmentation prediction. Wang et al.~\cite{wang2020cvpr} improve the generation of CAMs with a siamese network architecture and equivariant regularization to further improve results. Oquab et al.~\cite{oquab2015cvpr} use global max-pooling to produce image-level class predictions during training, where they use the activation maps for object localization. In this work, we choose a slightly different approach to previous methods by using random sampling instead of pooling to produce classification predictions. We show that this approach leads to better CAMs in terms of activating over the whole extent of objects as opposed to focusing on small discriminative regions. Therefore, it is well suited for the task of semantic segmentation. Lee et al.~\cite{lee2019cvpr} also use stochasticity for improving CAMs with spatial dropout for hidden unit selection in order to produce multiple localization maps during both training and inference. Our approach differs in two ways. First, we use a non-uniform sampling strategy based on the class activation maps for sampling image-level classification predictions. Second, we only use it during training, thus employing a deterministic inference scheme.

\textbf{Prediction-propagation using feature similarities.} The concept of propagating a signal to spatial neighbours based on feature similarities has been studied in the past, whether it be by computing a weighted mean of nearby pixels for image denoising~\cite{buades2011ipol}, or by letting the raw image features define the pair-wise potentials in a fully connected conditional random field (CRF) for propagating initial segmentation predictions~\cite{krahenbuhl2011neurips}. More recently it has been used in an attention-based WSSS approach where the CAM predictions are propagated to similar pixels based on deep features in order to improve segmentation results~\cite{wang2020cvpr}. In this work, we propose to do this implicitly through a \textit{feature similarity loss term} which is minimized when the prediction contours of CAMs match the edges in the image.

\textbf{Model distillation.} Several state-of-the-art WSSS methods~\cite{wang2020cvpr, lee2019cvpr, liu2020tpami} implicitly take advantage of model distillation by dividing up the framework into two stages, namely the generation of pseudo-labels using a classification network, and subsequently training a segmentation network on the generated pseudo-labels in a fully supervised setting. As has been demonstrated in the past~\cite{bucilua2006sigkdd, hinton2015arxiv}, large ensembles of models can be distilled into a single model without a significant loss of performance by training the final model to predict the output of the ensemble. Recently, it has also been shown that only one model is sufficient to be used as the ``ensemble'' for improved results compared to simply training the final model directly on the data~\cite{allen2020arxiv}. This is referred to as self-distillation. In a sense, the two-stage framework in WSSS is a form of self-distillation, and in this work we also take advantage of this phenomenon by adapting the two stages in generating pseudo-labels for the final segmentation network.

\section{Approach}

In this section we describe our approach of training a network for predicting CAMs, and how we integrate our two main contributions \textit{importance sampling} and \textit{feature similarity loss} for improving them.

\subsection{Computing CAMs}
\label{sec_computing_cams}

Let $a_\theta (x) \in [0, 1]^{W \times H \times K}$ denote a class activation map (CAM) which is a function of the input image~$x$, and parametrized by $\theta$, where $W$, $H$ and $K$ denote the width, height and number of classes (including background) respectively. Furthermore, let $s_\theta (x)_{ijk}$ represent the unnormalized logit of class $k$ at the position indexed by $i$ and~$j$. If we model the class probabilities in each pixel as a normalized probability distribution we can estimate the probability that a pixel contains a certain class $k$ using the softmax function as
\begin{equation}
	a_\theta (x)_{ijk} := \mathrm{Pr}(z_{ij} = k | x) = \frac{e^{s_\theta (x)_{ijk}}}{\sum_{t=1}^K e^{s_\theta (x)_{ijt}}},
\end{equation}
where $z \in \mathbb{R}^{W \times H}$ is the ground-truth segmentation mask and $z_{ij}$ is the class index present at the position indexed by $i$ and $j$. ${\mathrm{Pr}(z_{ij} = k | x)}$ is the estimated probability that class $k$ is present in the pixel indexed by $i$ and $j$ in a given image $x$. Since $z$ is unknown in the weakly supervised setting, the pixel-wise predictions need to be condensed into an image-level prediction. Commonly, global average pooling on the unnormalized logits or log-probabilities has been used in WSSS for this purpose~\cite{wang2020cvpr, ahn2018cvpr}. However, when minimizing a classification loss in this case, the global optimum occurs in theory when all pixels are predicted as belonging to some object class with high certainty. Therefore, we choose a different approach.

Assuming that, in order for an image to be classified as containing an object, it suffices that only one pixel contains that object, then we predict \mbox{$\mathrm{Pr} (k \in \{ z_{ij} \}_{i=1,j=1}^{W,H} | x)$}. We sample a single pixel per class based on $a_\theta$ since this essentially translates to ``if at least one pixel contains an object, the whole image contains this object''. The first obvious option is to apply global max pooling for this purpose, where the predicted image-level probabilities $\hat{y}$ can be written as
\begin{equation}
	\hat{y}_k := \mathrm{Pr} \left( k \in \{ z_{ij} \}_{i=1,j=1}^{W,H} \big{|} x \right) = \max_{ij} a_\theta (x)_{ijk}.
\end{equation}

The parameters $\theta$ can be found by minimizing the sum of $K$ binary cross-entropy loss terms
\begin{equation}
	\mathcal{L}_{\mathrm{ce}} (y, \hat{y}; \theta) = -\frac{1}{K} \sum_{k=1}^K y_k \log \hat{y}_k + (1 - y_k) \log (1 - \hat{y}_k),
\end{equation}
where $y_k$ is the image-level label for class $k$, which is equal to 1 if class $k$ is present in the image and 0 otherwise.

A shortcoming of max pooling is that it tends to activate over small discriminative regions and does not offer very useful segmentation predictions, even in cases where the classification prediction is correct. This can be observed in Figure~\ref{fig_comparison}. To improve the CAMs in this regard we use \textit{importance sampling} to produce stochastic image-level class predictions during training.

\subsection{Importance sampling}
\label{sec_importance_sampling}

As described in the previous section, global max pooling tends to activate over small discriminative regions and does not leverage adequate segmentation predictions for pseudo-label generation. To solve this problem we introduce an additional image-level prediction by sampling one pixel for each class using the probability mass function induced by the class activation map $a_\theta$. Let us define $K$ probability mass functions, one for each class
\begin{equation}
	\label{eq_importance_sampling_pmf}
	p_k(i, j | x ) = \mathrm{Pr} (I=i, J=j | x, k) = Z_k(a)^{-1} a_\theta(x)_{ijk},
\end{equation}
where $Z_k(a) = \sum_{i=1}^W \sum_{j=1}^H a_\theta(x)_{ijk}$ is a normalizing constant. Now, we sample image coordinates for each class which we use to extract the class activations. These activations are then interpreted as classification predictions $\tilde{y}_k = a_\theta(x)_{ijk}$, where $(i, j) \sim p_k$.

We train our CAM network using a classification loss containing two cross-entropy terms, one for the prediction $\hat{y}$ computed using max pooling, and one for the prediction $\tilde{y}$ attained from random sampling. Our classification loss term is a convex combination of these terms\pagebreak
\begin{equation}
	\label{eq_cls_loss}
	\mathcal{L}_{\mathrm{cls}}(y, \hat{y}, \tilde{y}) = (1 - \lambda) \mathcal{L}_{\mathrm{ce}} (y, \hat{y}) + \lambda \mathcal{L}_{\mathrm{ce}} (y, \tilde{y}),
\end{equation}
where $\lambda \in [0, 1]$ is a parameter controlling the weight between the two terms. A value of $0$ corresponds to the case in Section~\ref{sec_computing_cams} without importance sampling, and a value of $1$ corresponds to only using stochastic predictions. While the first term has been successful in classification tasks, it is unclear whether the first, second or a combination of the two is most suitable for weakly-supervised segmentation.

In our early experiments we observed that importance sampling improved the CAMs in terms of covering a larger extent of objects, but the prediction borders did not align with their edges, see Figure~\ref{fig_comparison}. For this reason we introduce a \textit{feature similarity loss term} which aims to match the prediction contours with the edges of objects.

\subsection{Feature similarity loss}
\label{sec_feature_similarity_loss}

Intuitively, similar pixels that are in close proximity have a high probability of being part of the same object. Additionally, if two nearby pixels are dissimilar, there is a chance that they belong to different classes and that the contour runs somewhere between them. Based on this, we formulate a loss term which penalizes dissimilar predictions for nearby similar pixels. Furthermore, similar predictions are discouraged for nearby dissimilar pixels if their predictions are sufficiently dissimilar to begin with. We also find it beneficial to learn the cross-over point of what should be considered as sufficiently dissimilar. In what follows, we formulate our feature similarity loss term as a function of the pixel-wise class predictions and features. Subsequently, we describe the intuition behind our formulation.

In the following equations we use a single index for the image coordinates to reduce notational clutter. Let us define a gating function $g(a_i, a_j) : \mathbb{R}^{2K} \rightarrow \mathbb{R}_+$ between the predictions $a_i$ and $a_j$ of the pixels $i$ and $j$, as well as a function $f( \delta ) : [0, 1] \rightarrow [-1, 1]$ which maps the dissimilarity $\delta(x_i, x_j) \in [0, 1]$ between the features $x_i$ and $x_j$ monotonically to $[-1, 1]$. Formulate the feature similarity~loss~term
\begin{equation}
	\label{eq_fsl_loss}
	\mathcal{L}_{\mathrm{fs}}(a, x) = -(HW)^{-2}{\textstyle\sum}_{ij} w_{ij} \, g(a_i, a_j) \, f(x_i, x_j),
\end{equation}
where $w_{ij}$ is a spatial weight defined using a Gaussian neighbourhood
\begin{equation}
	\label{eq_gauss_weights}
	w_{ij} = (2 \pi \sigma^2)^{-1} \text{exp} \left( -\left\lVert p_i - p_j \right\rVert^2_2 / (2 \sigma^2) \right),
\end{equation}
where $p_i$ is a two-dimensional vector containing the image coordinates of pixel $i$, and $\sigma$ is the standard deviation. Note that $a_i$ and $x_i$ are vectors representing respectively the class probability distribution and image features for pixel $i$. Furthermore, define the gating function using the $L^2$ distance ${g(a_i, a_j) = \frac{1}{2} \lVert a_i - a_j \rVert_2^2}$, and $f$ as
\begin{equation}
	\label{eq_fsl_f1}
	f(\delta(x_i, x_j)) = \text{tanh} \left( \mu + \text{log} \left( \delta / (1 - \delta) \right) \right),
\end{equation}
where $\mu$ is a bias parameter. The logarithm in \eqref{eq_fsl_f1} computes the logit of the binary decision problem whether or not two pixels are (dis)similar. The $\tanh$ function then maps the logit with the added bias to $[-1, 1]$. Thus, $f$ takes the values $-1$ and $+1$ when pixels are similar ($\delta=0$) and dissimilar ($\delta=1$) respectively, and $\mu$ controls the cross-over point at which~$f$ goes from negative to positive.

For two similar pixels $i$ and $j$ we have $f < 0$ and get $\mathcal{L}_{\mathrm{fs}} \geq 0$ since $g \geq 0$. $\mathcal{L}_{\mathrm{fs}}$ is thus minimized if $g$ is minimized, i.e.~if $a_i = a_j$. In the case of two dissimilar pixels on the other hand, i.e.~if $f > 0$, we have $\mathcal{L}_{\mathrm{fs}} \leq 0$, which is minimized if $g$ is maximized. This occurs when $a_i$~and~$a_j$ are opposite predictions, i.e.~if they are two one-hot vectors predicting different classes. But it is not always the case that two dissimilar pixels are part of different classes. For example, an object could contain some high-frequency texture or be made up of several parts with different visual appearance. However, since the gating function is chosen to be the squared $L^2$ distance, the gradient of $\mathcal{L}_{\mathrm{fs}}$ with respect to the predictions is proportional to the difference between the predictions, and equal to zero if $a_i = a_j$. Consequently, if two pixels have similar predictions, only a small gradient is propagated through the feature similarity loss, and the total gradient is dominated by the classification loss. This allows the network to classify larger regions to the same class, even though they contain parts with dissimilar features. Thus, the network can activate over the whole extent of objects.

Instead of manually finding good values for the standard deviation~$\sigma$ in \eqref{eq_gauss_weights} and the bias parameter~$\mu$ in \eqref{eq_fsl_f1}, we learn them together with the network parameters~$\theta$ during training. Furthermore, we use RGB pixel values in $[0, 1]$ for the features $x$, and choose the dissimilarity $\delta = \lVert x_i - x_j \rVert_1 / C$, which has been used in stereo matching~\cite{galar2013oe}, where $C=3$ is the dimensionality of the feature vectors. Although learnable features would allow for finding image-specific biases, they could potentially lead to trivial solutions or unwanted behaviours if combined with the classification loss, as we would essentially be learning the loss function. Therefore, we stick to RGB features.

\section{Experiments}

This section outlines the experimental results achieved for the methods described above, including implementation details, an ablation study, a baseline comparison, and a state-of-the-art comparison.

\subsection{Implementation details}
\label{sec_implementation_details}

For training and evaluation we use the \textit{PASCAL Visual Object Classes} (VOC) dataset~\cite{everingham2010ijvc} containing 1,464 training images, 1,449 validation images and 1,456 hold-out test images, with 20 foreground classes. Furthermore, we include the data presented by Hariharan et al.~\cite{hariharan2011iccv} which is common in VOC experiments, resulting in a total of 10,582 training images. Note, we only use image-level labels for training.

We use SEAM~\cite{wang2020cvpr} as a baseline and train our CAM network in a similar fashion, substituting global average pooling with importance sampling, and use the loss terms in~\eqref{eq_cls_loss} and~\eqref{eq_fsl_loss}. For a fair comparison, we also train an AffinityNet~\cite{ahn2018cvpr} to further refine our CAMs before pseudo-label generation. We modify the background parameter $\alpha$ to 2~and~4 when amplifying and weakening the background activations respectively. This was necessary as the values in our CAMs were distributed close to either 0 or 1, while the CAMs from SEAM were distributed evenly over $[0, 1]$. Otherwise, we use the same hyperparameters. Lastly, we train a DeepLab-v1~\cite{chen2015iclr} as our final segmentation model supervised by our pseudo-labels, and use CRF~\cite{krahenbuhl2011neurips} during inference. All networks use the same ResNet-38~\cite{wu2019pr} backbone. In our experiments we use two V100 32GB GPUs.

For evaluation, we use the commonly adopted mean intersection-over-union (mIoU) metric based on the area of the segmentation masks to measure region similarity. For a complementary view, we also use the F-score computed on the contours of the segmentation masks, which has previously been used to measure contour accuracy in video object segmentation~\cite{perazzi2016cvpr}. We adapt the code from Perazzi et al.~\cite{perazzi2016cvpr} and compute the F-score by first accumulating bipartite matches of the boundaries between the predicted and ground-truth masks, efficiently approximated using morphological operations. Subsequently, we compute the class-wise F-scores over the dataset which we then average. Although the VOC segmentation masks contain a thin no-class region between objects, the contour accuracy still gives a clear indication of the contour quality since we match the boundaries using dilated regions. Thus, we compute the contour accuracy by considering this no-class region as background.

\subsection{Ablation study}
\label{sec_ablation_study}

\begin{figure}[p]
	\setlength\abovecaptionskip{-1pt}
	\centering
	\includegraphics[width=\linewidth]{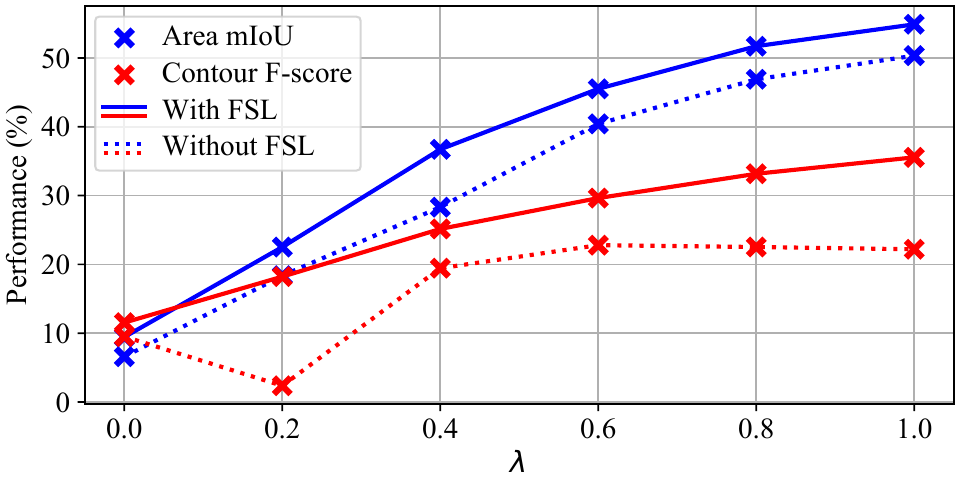}
	\captionof{figure}{Performance as a function of the loss parameter~$\lambda$ and the feature similarity loss (FSL) on the VOC \textit{validation} set.}
	\label{fig_ablation}
\end{figure}

To investigate the effects of our contributions, we compute the region similarity and contour accuracy of the pseudo-labels generated from our CAM network on the VOC validation set. We do not train an AffinityNet nor a DeepLab-v1 segmentation network, and neither do we use CRF at this stage. In Figure~\ref{fig_ablation} we sweep the loss parameter $\lambda$ from \eqref{eq_cls_loss}, and plot the area mIoU and contour F-score. We observe a significant increase in segmentation performance when increasing~$\lambda$ for both metrics. Thus, we have answered our question posed in Section~\ref{sec_importance_sampling} whether max pooling, importance sampling or a combination of the two is more suitable for weakly-supervised segmentation, where the answer is importance sampling. In the rest of our experiments, we use a value of $\lambda=1$ and ignore max pooling. Additionally, the feature similarity loss term always yields a performance boost, most notable for the contour accuracy at large values of~$\lambda$.

\begin{figure}[p]
	\setlength\abovecaptionskip{-8pt}
	\begin{center}
	\addtolength{\leftskip} {-2cm}
    \addtolength{\rightskip}{-2cm}
	\begin{tabular}{c>{\hspace{-10pt}}c>{\hspace{-10pt}}c>{\hspace{-10pt}}c>{\hspace{-10pt}}c}
		\includegraphics[width=.195\linewidth]{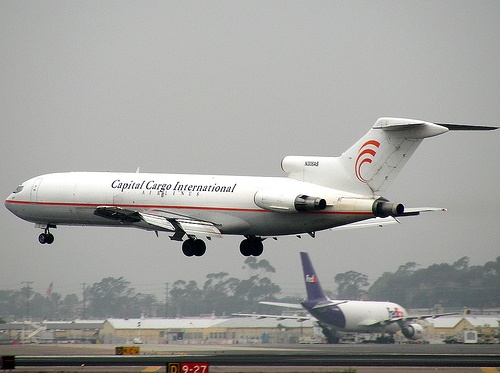} &
		\includegraphics[width=.195\linewidth]{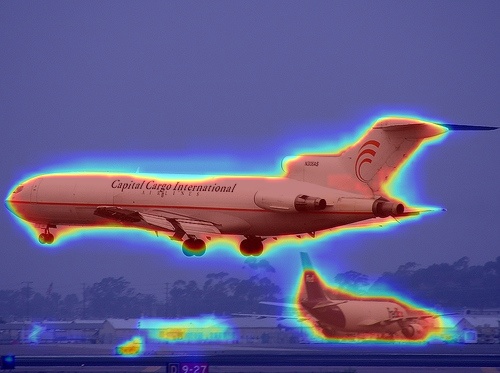} &
		\includegraphics[width=.195\linewidth]{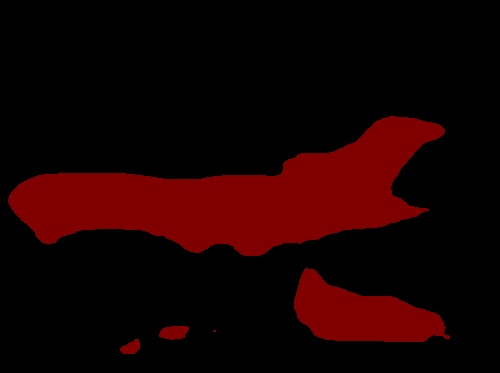} &
		\includegraphics[width=.195\linewidth]{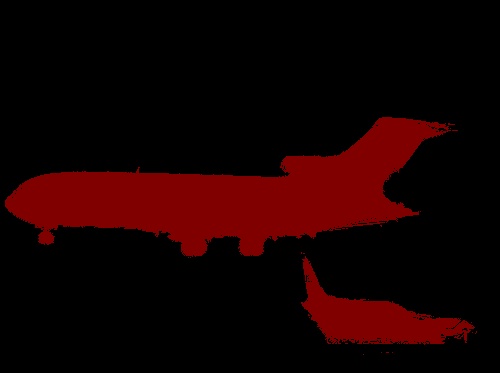} &
		\includegraphics[width=.195\linewidth]{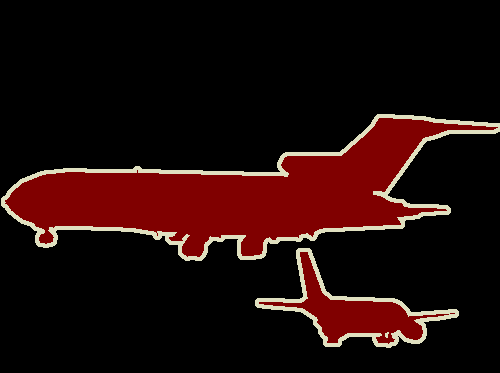} \\[-1pt]
		\includegraphics[width=.195\linewidth]{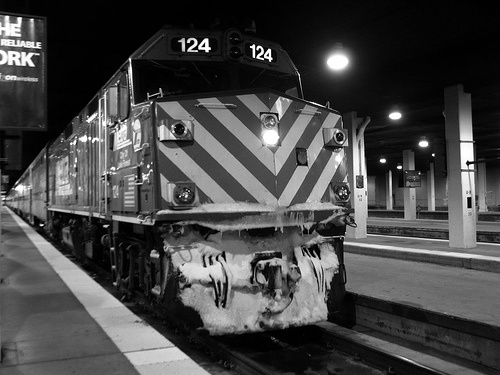} &
		\includegraphics[width=.195\linewidth]{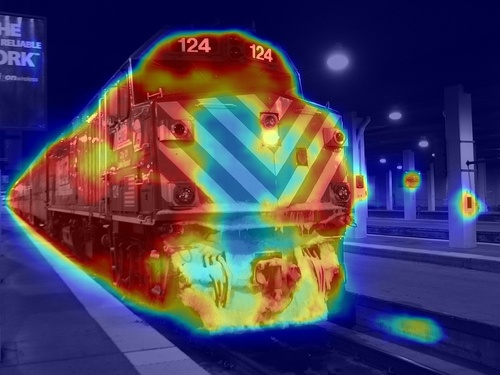} &
		\includegraphics[width=.195\linewidth]{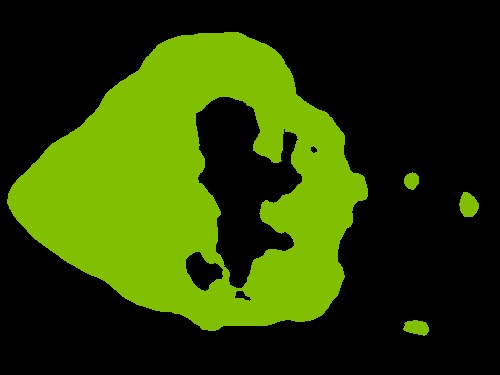} &
		\includegraphics[width=.195\linewidth]{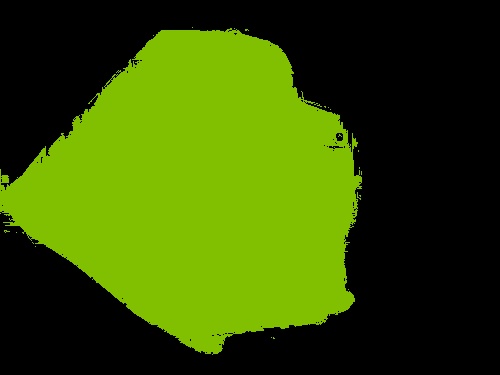} &
		\includegraphics[width=.195\linewidth]{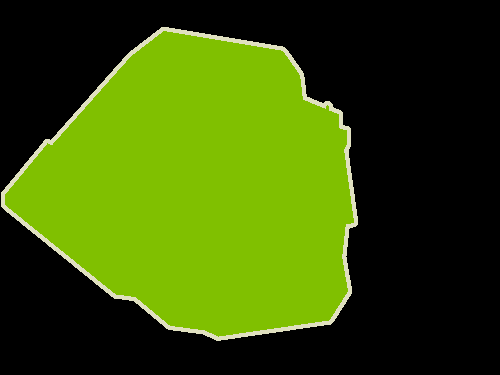} \\[-1pt]
		\includegraphics[width=.195\linewidth]{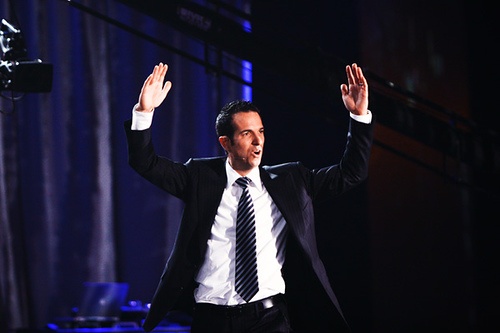} &
		\includegraphics[width=.195\linewidth]{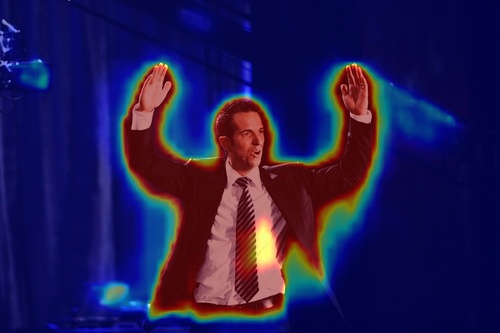} &
		\includegraphics[width=.195\linewidth]{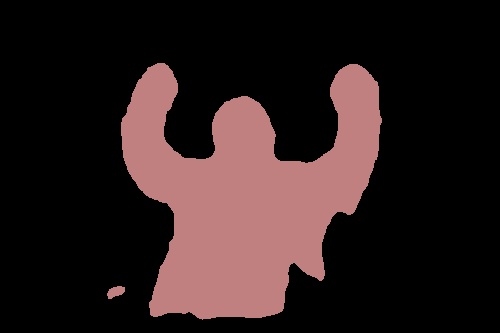} &
		\includegraphics[width=.195\linewidth]{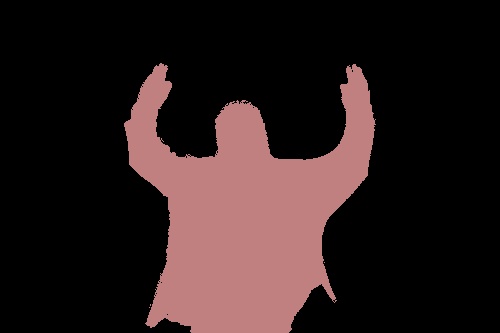} &
		\includegraphics[width=.195\linewidth]{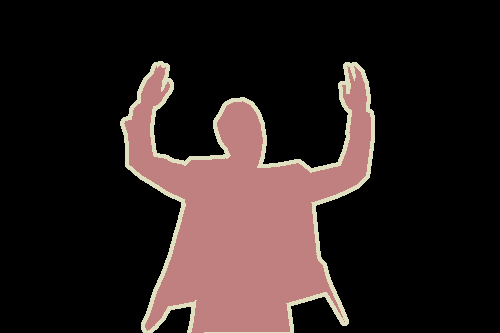} \\[-1pt]
		(a) & (b) & (c) & (d) & (e)
	\end{tabular}
	\end{center}
	\caption{Qualitative segmentation results. (a) Input image, (b) foreground class activations, (c) pseudo-labels from our CAM network, (d) final segmentation predictions and (e) ground-truth segmentations.}
	\label{fig_qualitative_results}
\end{figure}

\begin{table}[p]
	\setlength\abovecaptionskip{0pt}
	\centering
	\caption{Baseline comparison with SEAM~\cite{wang2020cvpr} on the VOC \textit{validation} set. Best results in bold for each comparison. The contour F-scores for SEAM have been computed by us.}
	\begin{tabular}{lcccc}
		\toprule
		& \multicolumn{2}{c}{Area mIoU} & \multicolumn{2}{c}{Contour F-score} \\
		\cmidrule(r){2-3}
		\cmidrule(r){4-5}
		& SEAM & Ours & SEAM & Ours \\
		\midrule
		CAMs          & 52.5 & \textbf{54.9} & 20.8 & \textbf{35.6} \\
		+ AffinityNet & 60.1 & \textbf{62.2} & 35.7 & \textbf{44.7} \\
		+ DeepLab-v1  & 64.5 & \textbf{64.6} & 38.2 & \textbf{43.8} \\
		\bottomrule
	\end{tabular}
	\label{table_baseline_comparison}
\end{table}

\begin{table}[p]
	\setlength\abovecaptionskip{0pt}
	\centering
	\caption{Area mIoU comparison on the VOC dataset.}
	\begin{tabular}[t]{llcc}
		\toprule
		Method                             & Backbone   & \textit{val} & \textit{test} \\
		\midrule
		CCNN~\cite{pathak2015iccv}         & VGG-16~\cite{simonyan2015iclr} & 35.3         & 35.6          \\
		SEC~\cite{kolesnikov2016eccv}      & VGG-16~\cite{simonyan2015iclr} & 50.7         & 51.1          \\
		STC~\cite{wei2016tpami}            & VGG-16~\cite{simonyan2015iclr} & 49.8         & 51.2          \\
		MDC~\cite{wei2018cvpr}             & VGG-16~\cite{simonyan2015iclr} & 60.4         & 60.8          \\
		MCOF~\cite{wang2018cvpr}           & ResNet-101~\cite{he2016cvpr}   & 60.3         & 61.2          \\
		SeeNet~\cite{hou2018neurips}       & ResNet-101~\cite{he2016cvpr}   & 63.1         & 62.8          \\
		AffinityNet~\cite{ahn2018cvpr}     & ResNet-38~\cite{wu2019pr}      & 61.7         & 63.7          \\
		IRNet~\cite{ahn2019cvpr}           & ResNet-50~\cite{he2016cvpr}    & 63.5         & 64.8          \\
		FickleNet~\cite{lee2019cvpr}       & ResNet-101~\cite{he2016cvpr}   & 64.9         & 65.3          \\
		SSDD~\cite{shimoda2019iccv}        & ResNet-38~\cite{wu2019pr}      & 64.9         & 65.5          \\
		SEAM~\cite{wang2020cvpr}           & ResNet-38~\cite{wu2019pr}      & 64.5         & 65.7          \\
		OAA~\cite{jiang2019iccv}           & ResNet-101~\cite{he2016cvpr}   & 65.2         & 66.4          \\
		CONTA~\cite{zhang2020neurips}      & ResNet-38~\cite{wu2019pr}      & 66.1         & 66.7          \\
		LIID~\cite{liu2020tpami}           & ResNet-101~\cite{he2016cvpr}   & 66.5         & 67.5          \\
		\midrule
		\textbf{Ours}                      & ResNet-38~\cite{wu2019pr}      & 64.6         & 65.2          \\
		\bottomrule
	\end{tabular}
	\label{table_sota}
\end{table}

\subsection{Baseline comparison}
\label{sec_baseline_comparison}

We now train an AffinityNet~\cite{ahn2018cvpr} on the CAMs generated by our network to further improve their quality. Subsequently, we train our final DeepLab-v1~\cite{chen2015iclr} segmentation model and evaluate its performance. For a qualitative assessment, we display in Figure~\ref{fig_qualitative_results} the foreground class activations, together with the pseudo-labels generated by our CAM network. Here, we also show the final segmentation predictions from the DeepLab-v1 model after CRF~\cite{krahenbuhl2011neurips} has been applied.

In Table~\ref{table_baseline_comparison} we compare region similarity and contour accuracy with our baseline SEAM~\cite{wang2020cvpr} at the three training stages. Note that the first two rows comparing CAM and AffinityNet results show the metrics computed for pseudo-labels without CRF. The final row compares the results between the final segmentation predictions where CRF has been applied. We observe that our method has significantly better contour accuracy across the board, where the difference is most notable for the CAM results. Furthermore, the property of accurate contours carries through to the final segmentation model. Both methods benefit from training an AffinityNet where the contour accuracy and region similarity increase for both methods.

\subsection{State-of-the-art comparison}
\label{sec_sota_comparison}

To compare our results with previous methods we evaluate the predictions of our DeepLab-v1 model on the test set. We submit our predictions to the PASCAL VOC evaluation server exactly once, and present the results in Table~\ref{table_sota}. We see that our method reaches similar performance in terms of region similarity compared to current state-of-the-art methods. Although LIID~\cite{liu2020tpami} also perform experiments using the more powerful Res2Net-101~\cite{gao2021tpami} backbone, we choose to compare with their performance using the ResNet-101~\cite{he2016cvpr} backbone in Table~\ref{table_sota} as it is more comparable to the backbone architectures used in our method as well as in the other entries of the table.

\section{Conclusions}
\label{sec_conclusions}

In this paper, we present two methods for improving segmentation predictions in weakly-supervised semantic segmentation. First, we use importance sampling for producing image-level classification predictions during training, which results in CAMs that activate over a larger extent of objects. Second, we propose a new loss term for matching prediction contours with edges in the image. We show experimentally that this significantly improves the contour accuracy of our segmentation predictions. Finally, we perform experiments on the PASCAL VOC 2012 benchmark dataset to show that our method is comparable to state-of-the-art methods in terms of region similarity.

\newpage

\bibliographystyle{IEEEbib_abbrv}
\bibliography{refs}

\end{document}